\documentclass[conference]{IEEEtran}
\IEEEoverridecommandlockouts
\usepackage{cite}
\usepackage{amsmath,amssymb,amsfonts}
\usepackage{algorithmic}
\usepackage{graphicx}
\usepackage{textcomp}
\usepackage{xcolor}
\usepackage{hyperref}
\usepackage[frozencache]{minted}

\usepackage{caption}
\def\BibTeX{{\rm B\kern-.05em{\sc i\kern-.025em b}\kern-.08em
    T\kern-.1667em\lower.7ex\hbox{E}\kern-.125emX}}
\begin{document}

\title{Lodestar: An Integrated Embedded Real-Time Control Engine\\
\thanks{Research reported in this publication was supported by the National Institute of Biomedical Imaging and Bioengineering of the National Institutes of Health under award number R01EB029766, as well as the National Aeronautics and Space Administration under award number 80NSSC21K1030.}
}

\author{\IEEEauthorblockN{Hamza El-Kebir}
\IEEEauthorblockA{\textit{Dept. of Aerospace Engr.} \\
\textit{University of Illinois}\\
\textit{Urbana-Champaign}\\
Urbana, IL 61801, USA \\
\texttt{elkebir2@illinois.edu}}
\and
\IEEEauthorblockN{Joseph Bentsman}
\IEEEauthorblockA{\textit{Dept. of Mechanical Sci. \& Engr.} \\
\textit{University of Illinois}\\
\textit{Urbana-Champaign}\\
Urbana, IL 61801, USA \\
\texttt{jbentsma@illinois.edu}}
\and
\IEEEauthorblockN{Melkior Ornik}
\IEEEauthorblockA{\textit{Dept. of Aerospace Engr.,} \\
\textit{Coordinated Science Lab.} \\
\textit{University of Illinois}\\
\textit{Urbana-Champaign}\\
Urbana, IL 61801, USA \\
\texttt{mornik@illinois.edu}}
}

\maketitle

\begin{abstract}
In this work we present \textit{Lodestar}, an integrated engine for rapid real-time control system development. Using a functional block diagram paradigm, Lodestar allows for complex multi-disciplinary control software design, while automatically resolving execution order, circular data-dependencies, and networking. In particular, Lodestar presents a unified set of control, signal processing, and computer vision routines to users, which may be interfaced with external hardware and software packages using interoperable user-defined wrappers. Lodestar allows for user-defined block diagrams to be directly executed, or for them to be translated to overhead-free source code for integration in other programs. We demonstrate how our framework departs from approaches used in state-of-the-art simulation frameworks to enable real-time performance, and compare its capabilities to existing solutions in the realm of control software. To demonstrate the utility of Lodestar in real-time control systems design, we have applied Lodestar to implement two real-time torque-based controller for a robotic arm. In addition, we have developed a novel autofocus algorithm for use in thermography-based localization and parameter estimation in electrosurgery and other areas of robot-assisted surgery. We compare our algorithm design approach in Lodestar to a classical ground-up approach, showing that Lodestar considerably eases the design process. We also show how Lodestar can seamlessly interface with existing simulation and networking framework in a number of simulation examples.
\end{abstract}


\section{Introduction}

Cyber-physical systems often exhibit complex dynamic behaviors that necessitate the adoption of advanced control laws and their safe and efficient implementation. Many of these systems lie at the intersection of a variety of disciplines, with their control systems often requiring custom software and hardware solutions tailored to the specific application. As an example, one may consider autonomous vehicles, which require an understanding of the drivetrain systems, vision systems and computer vision algorithms for parsing the surroundings, and control and decision algorithms to guide the vehicle \cite{Kato2015}. In this application alone, one finds a need to incorporate low-level control algorithms for the drivetrain, as well as complex computer vision algorithms for image segmentation, vehicle classification and localization, and a high-level decision algorithm that ultimately guides the vehicle \cite{Madni2018}. Another example, which is pertinent in the latter part of this work, lies in robot-assisted surgery, specifically in relation to \emph{electrosurgery}. Here, one aims to provide real-time diagnostics to surgeons, a challenge that lies at the intersection of computer vision, signal processing, and control \cite{El-Kebir2021d}, while also requiring the capability of passing sensitive information in a secure manner between networked systems to safeguard the patient's medical information.

\begin{figure}[t]
\centering
\includegraphics[width=\linewidth]{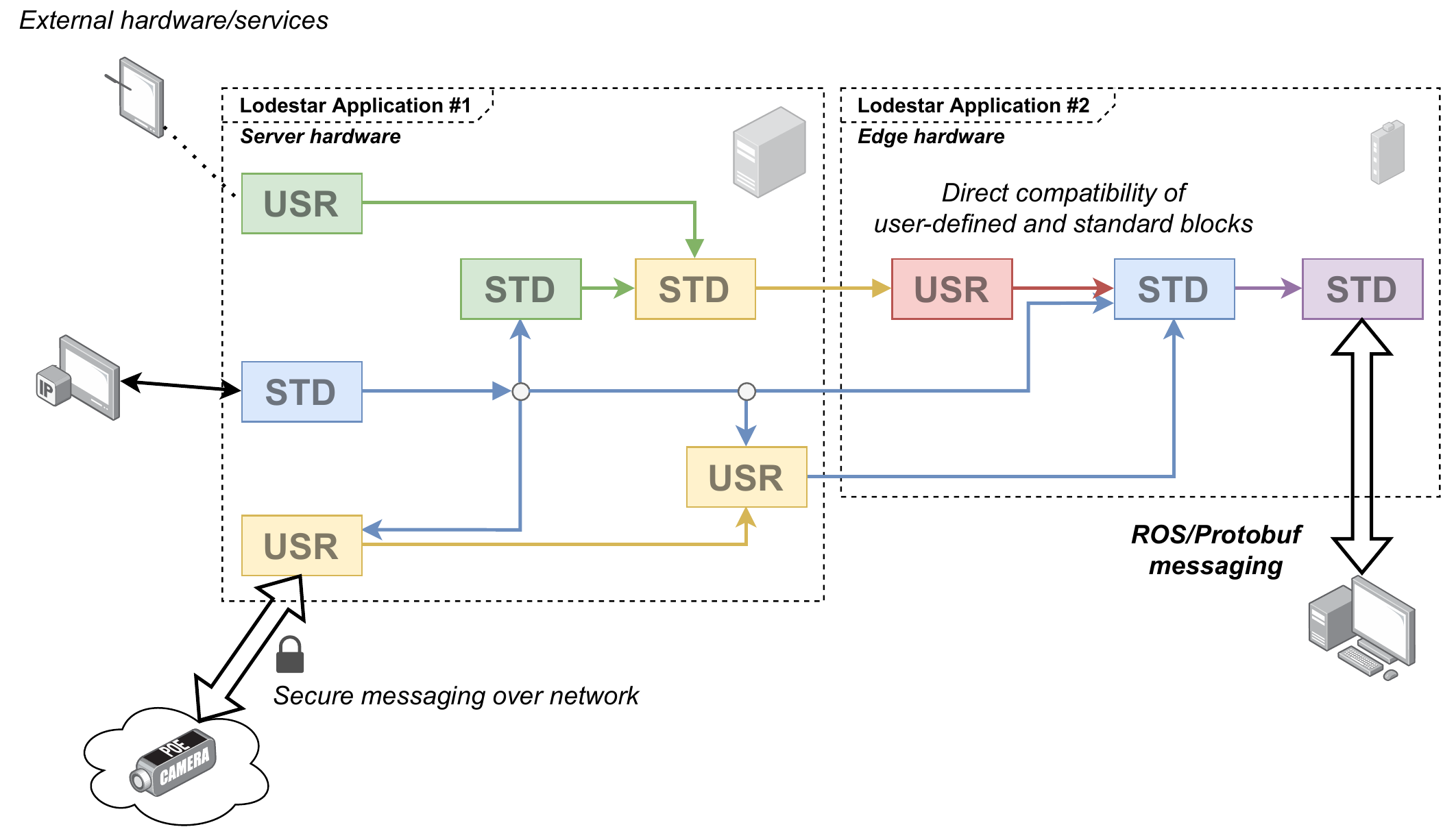}
\caption{Overview of the Lodestar engine. A description of the complete system is provided by the user, with internal execution order and data dependencies being resolved automatically. External hardware and software services can easily be interfaced with, using existing protocols or using a custom encrypted communication protocol. Lodestar directly provides a baseline implementation that is lightweight and efficient enough to execute directly on target hardware, without modifications or significant overhead.}
\label{fig:lodestar-overview}
\end{figure}

Both in industry and academia, such problems are almost exclusively solved using custom-built solutions, with little standardization insofar as software is concerned. While solutions such as Simulink present themselves as \emph{de facto} industry standards \cite{Rajhans2018}, proprietary software interfaces and a lack of portability often make it hard to obtain immediately executable code from these frameworks. Moreover, during prototyping, it is highly likely that some of the hardware or software used in a system does not have predefined interfaces with control design software, necessitating the need for custom interfaces and wrappers. In practice, one would therefore either need to call on a team of engineers proficient in each of the subfields and their associated software stacks, or become proficient in all of the required software packages themself \cite{El-khoury2005}. Recognizing this major hurdle in rapid control software prototyping and deployment, the objective of this work is to introduce a new open-source control engine, called \emph{Lodestar}\footnote{\href{https://ldstr.dev}{https://ldstr.dev}}, that aims to address issues in hardware and software abstractions while still providing real-time performance levels. Lodestar addresses two key challenges: (i) enabling user-extensible abstraction of domain-specific hardware and software functionalities as part of a functional block diagram description of a system, and (ii) producing directly executable code for rapid prototyping, including networking capabilities. Lodestar is implemented in ISO C++11 to maximize platform support, and is licensed under the permissive BSD 3-clause license\footnote{\href{https://github.com/helkebir/Lodestar/blob/master/LICENSE}{https://github.com/helkebir/Lodestar/blob/master/LICENSE}}. Fig.~\ref{fig:lodestar-overview} shows an overview of the Lodestar engine.

This paper is structured as follows. Related work is presented in Sec.~\ref{sec:prior work}. In Sec.~\ref{sec:preliminaries}, we discuss basic concepts in Lodestar, as well as design decisions that underpin the engine.
 Then, in Sec.~\ref{sec:applications}, two applications are given: (i) an overview of real-time performance of Lodestar as applied to joint tracking on a robotic arm, (ii) an application of Lodestar to real-time autofocus of an infrared thermographer. Finally, in Sec.~\ref{sec:conclusion}, conclusions are drawn, and future work is presented.
 
\section{Related Work}\label{sec:prior work}

Over the past decade, the introduction of the Robot Operating System (ROS, \cite{Quigley2009}) has made a significant impact on the way experimental robotics is practiced, providing a unified ecosystem for message passing between software and hardware services. This has afforded many users and vendors alike to abstract away hardware specifics, instead providing ``services'' that operate on their inputs, and produce a desired effect/output as a result. Users and other services may then subscribe to specific topics in which these results are broadcast, and may broadcast their own messages on topics of their choosing. While this publisher/subscriber, or pub/sub, framework has been regarded as a gold standard within the field of experimental robotics, it leaves the entirety of the control system synthesis to the end-user. In practice, this has compelled users to develop one-off solutions to their control problems, leaving a single unified solution to be desired, except in the most common of cases (such as PID-based joint control using \texttt{ros\_control} \cite{Chitta2017}).

When considering robotics simulation frameworks, tools such as Drake \cite{drake}, MoveIt \cite{Chitta2012}, Gazebo \cite{Koenig2004}, and CARLA \cite{Dosovitskiy2017}, have shown to present a great number of features for testing cyber-physical control systems. Since these tools are all simulation-focused, they often apply nondeterministic algorithms such as implicit solvers, which are not feasible in real life unless custom schemes are adopted \cite{Arnold2007}. As an example, Drake provides implicit Euler, Bogacki--Shampine, and Runge--Kutta schemes\footnote{\href{https://drake.mit.edu/doxygen\_cxx/group\_\_integrators.html}{https://drake.mit.edu/doxygen\_cxx/group\_\_integrators.html}}, all of which require access to the underlying model state at future times \cite{Arnold2007}. Moreover, a high-fidelity system model often precludes direct migration from a simulation environment to a real-life application, especially when this model is actively used within the controller architecture \cite{Arnold2007}. Most importantly, these tools do not aim to provide users with a comprehensive modeling toolset to design control systems, but rather supply a simulation framework that accepts the final control signals. As an illustration of this fact, while Drake does provide routines for linear quadratic regulator synthesis and PID controllers \cite{drake}, it lacks a framework for modular controller design; the same observation can be made for CARLA \cite{Dosovitskiy2017}. This latter point underlines the need of a directly deployable, open-source control engine, in which users can synthesize controllers, test them in simulation, and directly apply them in real-life.

Turning our attention to controller synthesis frameworks, we find a number of specialized toolchains do indeed exist. Recently, NASA has introduced the F Prime framework, which aids in designing system specifications (templates) for flight software stacks \cite{Bocchino2018}. F Prime generates boilerplate code, and leaves the actual logic specification to the user. Such an approach is patently different from Lodestar; we explicitly aim for a design framework that produces directly executable code. Previously, MATLAB's Simulink\footnote{\href{https://www.mathworks.com/products/simulink.html}{https://www.mathworks.com/products/simulink.html}} and Stateflow\footnote{\href{https://www.mathworks.com/products/stateflow.html}{https://www.mathworks.com/products/stateflow.html}} modeling frameworks were used as a basis for direct code translation \cite{Hanselmann1999}; this functionality is now directly provided by their publisher in the form of C/C++ code generation \cite{Rajhans2018}. While Simulink does allow for user-authored custom blocks to be introduced, their input--output structure is limited to a single matrix-type input and output. More importantly, the input and output may not have a \emph{direct algebraic dependency}, a limitation that is not present in Lodestar as is discussed later. Finally, a Simulink model may not be executed on hardware directly without writing custom interfaces that get compiled to the proprietary MEX executable format or using slower MATLAB-based interfaces. In the context of rapid prototyping, any Simulink models would run in a Java virtual machine (JVM) instance, which may incur appreciable time delays when using relatively few MEX functions. Most importantly, direct prototyping in Simulink requires the platform to have a working installation of MATLAB and Simulink, which may quickly become prohibitive on (embedded) edge devices. Our proposed solution only requires an ISO C++11 compliant compiler to be provided. 

In the realm of embedded applications, solutions based on Lustre \cite{Caspi1987}, Esterel \cite{Berry1992}, Signal \cite{Benveniste1991}, Ada \cite{Taft2013}, and other languages are commonly employed. In particular, applications using Ada and the Lustre-based SCADE framework, as well as IBM's Rational Rhapsody framework, are particularly pervasive in safety-critical applications such as automotive and aerospace thanks to their strongly certifiable design and implementation \cite{Colaco2017}; the aforementioned tools are, however, rarely encountered in experimental robotics due to the strictness of these frameworks and languages. Recently, CoCoSim \cite{Bourbouh2020} has aimed to bridge this gap, presenting a framework that translates Simulink models to Lustre; being based on Simulink, however, it presents the same portability issues as mentioned in the previous paragraph.

In prior work on executable code generation in the context of control system design, most approaches do not allow for circular data-dependencies, or algebraic loops \cite{Kim2003}. In Simulink, algebraic loops may appear in simulation and are promptly solved, but when using code generation capabilities, these loops may not be present in the final system \cite{Benveniste2017}. Algebraic loops often arise in constrained control algorithms, such as hard-constrained model predictive control \cite{Syaichu-Rohman2003}, which are essential in practical applications \cite{Adegbege2021}. While a common workaround to algebraic loops is to introduce a small time delay in one of the components to `break the loop,' such an approach removes any analytical robustness guarantees, and may not be feasible under strict execution time constraints to start with \cite{Syaichu-Rohman2002}. Another simulation framework, CyPhySim \cite{Brooks2015}, allows for algebraic loop solving, but these capabilities are confined to simulating models. In this work we show how Lodestar automatically detects algebraic loops, and produces efficient application-specific solvers that may be directly deployed.



\section{Concepts \& Architecture}\label{sec:preliminaries}

In this section, we present several notions that are central in the design of the Lodestar control engine. We also briefly discuss how these concepts have been implemented in Lodestar, as well as how users are able to extend the capabilities of Lodestar.


\subsection{Functional Blocks}

\begin{figure}[t]
\centering
\includegraphics[width=0.6\linewidth]{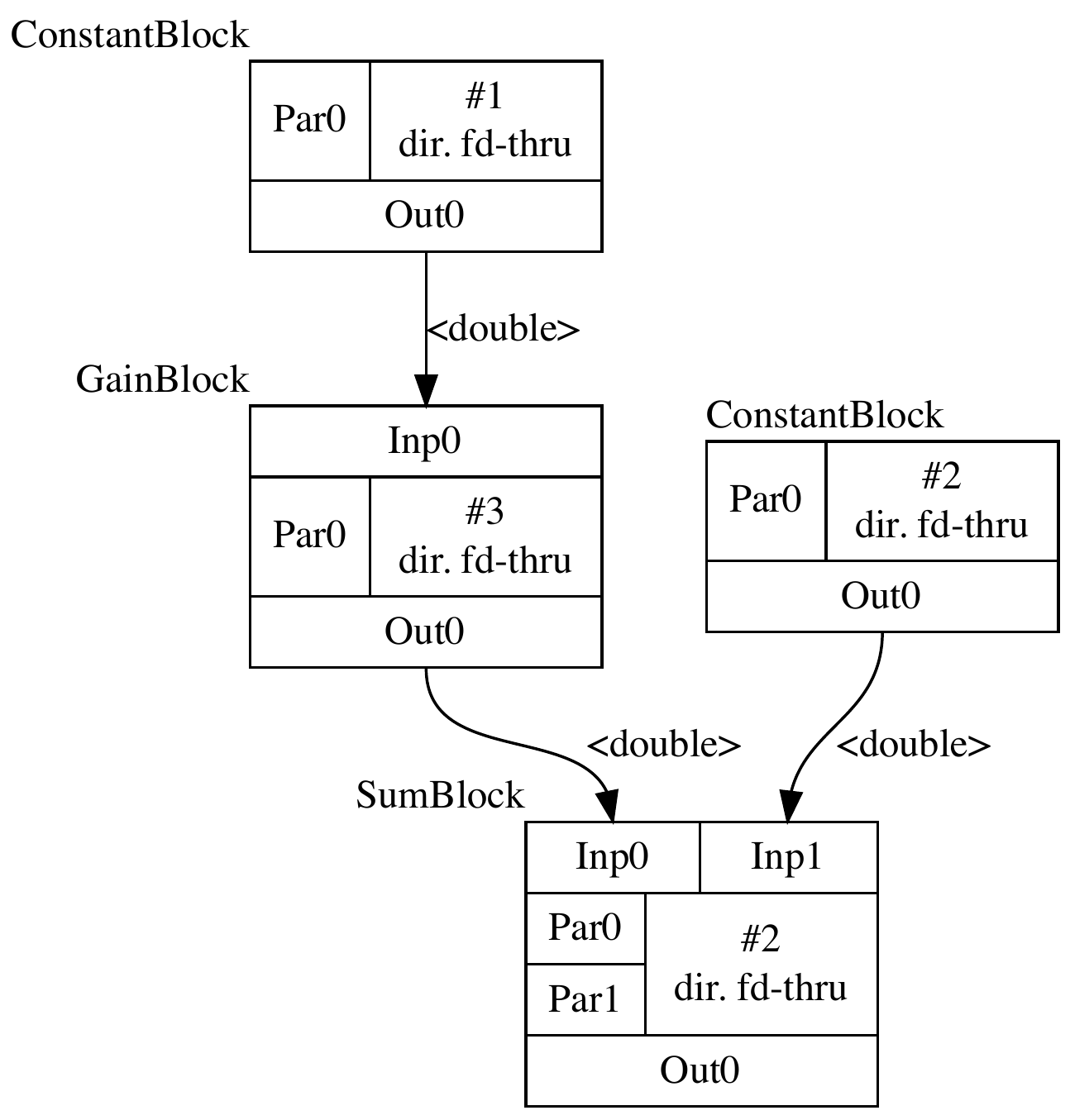}
\caption{Illustrative diagram of a system with complete feedthrough.}
\label{fig:simple-diagram}
\end{figure}

Functional blocks, or operators, have been extensively recognized as a natural way of modeling data-driven systems \cite{Wcislik2015}. In this work, we consider functional blocks that consist of zero or more inputs, outputs, and parameters (see Fig.~\ref{fig:simple-diagram}). In this context, a single input, output, or parameter may refer to a multi-dimensional object such as a matrix, or a more complex structure. Each functional block comes with a \emph{pure function}, i.e., a function modifying the block's internal state based on the inputs and parameters, exposing the results of this modification in its outputs.


In Lodestar, block inputs are only meant to be altered manually by a user prior to being engaged in an interconnection. Given that Lodestar is implemented in C++, compile-time checks are implemented to ensure that interconnections have the same types (i.e., connected input--output pairs must have the same type), among many other consistency checks (such as dimensions, multiplicability, etc.).

Pure functional blocks can also be viewed as atomic units of a control flow; each functional block provides well-defined behavior, which is naturally implementation-agnostic. This allows for direct code generation after the execution order is determined and circular dependencies are resolved. The resolution of execution order and circular dependencies is discussed in a later section.

We distinguish between two main classes of blocks: those that exhibit \emph{direct feedthrough}, and those data do not. The instantaneous output of a direct feedthrough block directly depends on the input (for example, a sum or product), while the instantaneous output of a block that does not have direct feedthrough may only depend on the inputs observed during the last cycle and before (e.g., an integrator or time delay block).

In Lodestar, a simple system of the form of Fig.~\ref{fig:simple-diagram} can be generated using the C++ code shown in Lst.~\ref{listing:simple block diagram}. In this example, we multiply the first constant by a gain, and sum it with the second constant, producing an output of 4.5 in this case. As can be seen, extensive use of template metaprogramming is made throughout Lodestar, providing specialized function implementations at zero overhead in many cases, as is described later. Lst.~\ref{listing:simple block diagram} shows how the blocks (\texttt{ConstantBlock}, \texttt{SumBlock}, \texttt{GainBlock}) each allow for their internal type to be defined, as well as the number of inputs in case of the \texttt{SumBlock}. A key advantage of Lodestar's design is the fact that many consistency checks may be performed at compile-time. Hidden in the implementation of the \texttt{GainBlock} is a static assertion that verifies whether the input is multiplicable with the gain, and the result of that multiplication can be cast to the expected output of the \texttt{GainBlock}. Such functionality is key when considering matrix-valued inputs.
\begin{listing}[!ht]
\begin{minted}[xleftmargin=6mm, frame=lines, linenos, autogobble, fontsize=\footnotesize, framesep=2mm]{cpp}
ConstantBlock<double> c{5}, c2{2};
SumBlock<double, 2> s;
GainBlock<double> g{0.5};
    
connect(c.o<0>(), g.i<0>());
connect(g.o<0>(), s.i<0>());
connect(c2.o<0>(), s.i<1>());

BlockPack bp{c, c2, s, g};
aux::Executor ex{bp};
ex.resolveExecutionOrder();
ex.trigger();
\end{minted}
\caption{A simple block diagram in Lodestar}
\label{listing:simple block diagram}
\end{listing}

\subsection{Interconnections and Execution Order}

In Lst.~\ref{listing:simple block diagram}, one can see how blocks are declared (lines 1--3). Each block inherits from a \texttt{BlockProto} object, which provides a virtual \texttt{trigger} function, as well as a unique block identifier. The \texttt{trigger} function is the only level of run-time polymorphism that Lodestar requires, making it very memory efficient \cite{Nath2012}.

In the Lst.~\ref{listing:simple block diagram}, it can be seen how outputs and inputs are interconnected using the \texttt{connect} function (lines 5--7). Accessing inputs, outputs, and parameters is much the same as in Modelica; the syntax is \texttt{blk.h<s>()}, where \texttt{h} is \texttt{i}, \texttt{o}, or \texttt{p}, and \texttt{s} is the \emph{slot index}. Given the templated implementation of Lodestar, code does not compile if invalid indices are accessed, preventing out-of-range access at run-time. Inputs and outputs are wrapped into a lightweight \texttt{Signal<T>} object, where each signal tracks its own interconnections, as well as its parent block and slot index; this information is used when computing the execution order and detecting algebraic loops. At run-time, Lodestar verifies that only outputs are connected to inputs, and that each input is connected to at most one output. In this way, consistency of the system is ensured before any execution may take place. Parameters are meant to be set prior to executing the system and should remain constant while the system is executed.

Before computing the overall execution order, blocks are bundled in a \texttt{BlockPack} object, where each block is reduced to the information provided by the corresponding \texttt{BlockProto} object, after extracting block-specific information (e.g., whether a block has direct feedthrough). With this \texttt{BlockPack} object, we can then initialize and \texttt{Executor} object, which allows Lodestar to determine the execution order. The execution is determined by the following rules \cite{Kim2003}:
\begin{enumerate}
    \item In case a block with direct feedthrough is driven by another block, the latter block takes precedence;
    \item Blocks with indirect feedthrough need not be ordered, other than by the first rule.
\end{enumerate}

These rules follow naturally when considering the composition of mathematical operator; the ordering of blocks with indirect feedthrough is left last since their output is not dependent on their current input. After ordering, the \texttt{Executor} object then provides its own \texttt{trigger} function, which performs one \emph{cycle} of the program in the correct order.

\subsection{Algebraic Loops and Circular Data Dependencies}

\begin{figure}[t]
\centering
\includegraphics[width=0.5\linewidth]{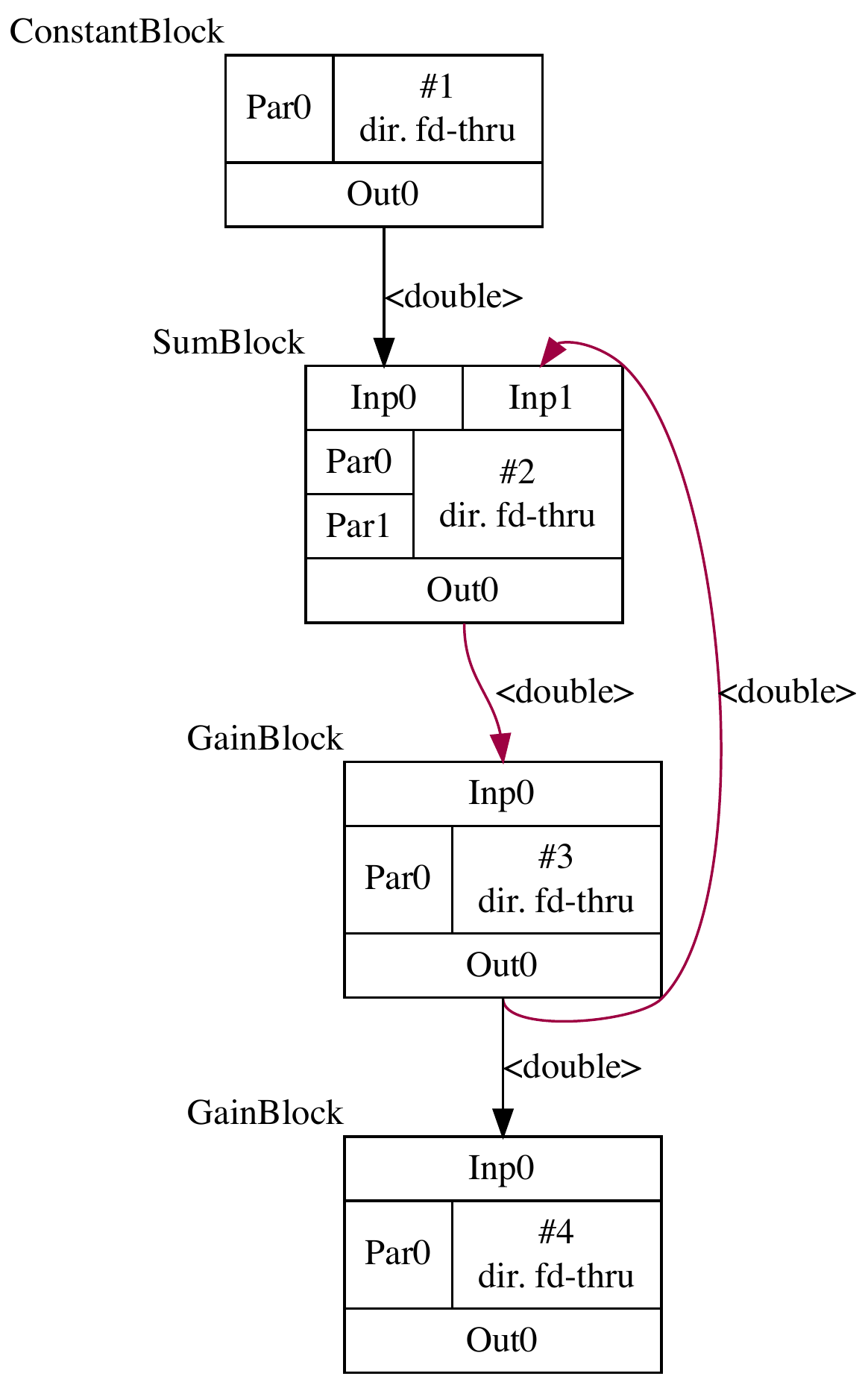}
\caption{Diagram of a system with an algebraic loop.}
\label{fig:algebraic-loop-diagram}
\end{figure}

Unlike most data-driven designs, which only consider systems where the block interconnections form a directed acyclic graph (DAG) \cite{Kim2003, Wcislik2015}, functional block models in Lodestar may contain cycles, or loops. Fig.~\ref{fig:algebraic-loop-diagram} shows a block diagram consisting of two constants, a gain, and a sum block. Marked in red is a circular data dependency, or \emph{algebraic loop}. Algebraic loops arise when cycles are formed that consist exclusively of blocks with direct feedthrough \cite{Syaichu-Rohman2003,Adegbege2021}. Commonly, these algebraic loops are `removed' by introducing indirect feedthrough blocks, such as time delays\footnote{\href{https://www.mathworks.com/help/simulink/ug/remove-algebraic-loops.html}{https://www.mathworks.com/help/simulink/ug/remove-algebraic-loops.html}} \cite{Syaichu-Rohman2002}. As mentioned in Sec.~\ref{sec:prior work}, such interventions may significantly alter the validity of the model, and may even cause systems to become unstable \cite{Syaichu-Rohman2002}.

As mentioned in the introduction, algebraic loops arise in many  applications, including constrained optimal control algorithms \cite{Syaichu-Rohman2003}, power electronics \cite{Castaings2014}, and adaptive control \cite{MaliniLamego2001}, as well as feedforward control \cite{Altmann2005}. To directly solve these systems of algebraic equations, Lodestar currently uses a symbolic algebra package GiNaC \cite{Bauer2002}. Direct feedthrough blocks are amenable to a symbolic (algebraic) description of their functionality, allowing us to extract each algebraic loop and expand each symbol in terms of known relations, external inputs, and block parameters. This expansion is performed by traversing the loop starting at each unknown variable $x_i$, until all variables have their own algebraic equation of the form $f_i (x) = 0$. Doing so yields $n$ equations with $n$ unknowns, where each of the $f_i$ is symbolically expressed. These symbolic expressions are then symbolically differentiated to obtain a Jacobian $\nabla f(x)$. A Newton--Raphson iteration is then constructed to solve for the variables:
\begin{equation*}
\begin{split}
    \nabla f(x^{(k)}) \Delta x^{(k)} &= -f(x^{k}), \\
    x^{(k+1)} &= x^{(k)} + \Delta x^{(k)}.
\end{split}
\end{equation*}
Lodestar is capable of generating the Newton--Raphson linear parametric systems, which may be solved using an efficient linear system solver coupled with an adequate stopping criterion. This departs from the approach used in Simulink, where the Jacobian is numerically approximated at the cost for an increased number of function evaluations \cite{Shampine1999}. While it is possible to turn to higher order solvers, this is often prohibitive when dealing with limited hardware; concerns regarding divergence of the scheme, while valid, cannot be addressed without resorting to global optimization schemes \cite{Shampine1999, Adegbege2021}. In practice, a sufficiently close initial choice would allow the algorithm to run reliably \cite{Shampine1999, Brooks2015}.

In the system in Fig.~\ref{fig:algebraic-loop-diagram}, blocks \#2 and \#3 are part of the algebraic loop, while output 0 of block \#1 forms an internal input to the loop. One can verify that the output of the sum block is $S = \frac{C}{1-G}$, where $C$ is the constant, and $G$ is the gain. The Newton--Raphson system will read $(G-1) \Delta S^{(k)} = -(C + G S^{(k)} - S^{(k)})$.

To the best of the authors' knowledge, Lodestar provides the first framework in which these algebraic equations are solved using application-specific solvers running in real-time (i.e., no general purpose nonlinear algebraic equation solvers).

%
%

\subsection{Control Routines}

Lodestar comes bundled with an extensively tested and documented standard library of blocks, currently counting over twenty blocks dealing with linear control systems, time delays, 2D convolution for images, PID controllers, etc. In addition to existing blocks, custom behavior can quickly be tested using the special \texttt{FunctionBlock}, in which users can easily author their own behavior. Lodestar also provides a number of commonly encountered control routines as standalone functions in C++, where these were, to the best of our knowledge, previously only available in proprietary software, interpreted languages such as Python, and Fortran (i.e., SLICOT \cite{Benner1999}). Currently, synthesis routines (e.g., discrete-time linear quadratic regulators), discretization algorithms (e.g., zero-order hold, bilinear transformations), and conversions between control systems (state-space to transfer functions and vice versa) are bundled together. Besides these capabilities, Lodestar also allows for symbolic systems to be defined, automatically generating C code that linearizes these systems. Combining these capabilities allows for expedient control system design and implementation, as demonstrated in Sec.~\ref{sec:applications}.

\subsection{Networking \& Message Passing}

When dealing with distributed systems, message passing often presents itself as one of the major challenges; this issue is exactly where ROS has found its calling \cite{Quigley2009}. However, ROS imposes stringent constraints on the project structure and the supported platforms, as well as the way in which messages are defined. This has led to the decision that Lodestar be independent of the ROS ecosystem as a baseline, but still allows for direct integration with ROS applications. A major goal for the networking aspect of Lodestar was that it should support as many platforms as possible, while remaining lightweight and flexible to mesh well with a functional block structure. To this end, Lodestar has adopted Google's Protocol Buffer\footnote{\href{https://developers.google.com/protocol-buffers}{https://developers.google.com/protocol-buffers}} framework. In particular, messages sent in Lodestar are always preceded by a \emph{herald} message that declares the origin/destination of the message, as well as information about the type of message. This allows Lodestar to discard messages that are wrongly addressed or malformed (e.g., a wrongly sized matrix). Therefore, the required buffer size is known at compile time, since the expected message types are determined by the blocks present in the system.

Most importantly, this layered approach to messaging allows us to encrypt sensitive messages, since their size is known a priori. The herald message therefore also includes a signature, which can be matched with a list of trusted public keys by the Lodestar program to determine whether a message should be trusted or not. Conversely, any messages that the Lodestar program sends are also encrypted by a private key with a corresponding public key that is known to the user (or client). This capability is not present in ROS \cite{Profanter2019}, making it ill-suited for privacy-sensitive applications such as the medical application considered in the next section, as well as long range communications with in extra-planetary mission simulators such as the NASA OceanWATERS platform \cite{Catanoso2021}.

%
%

%

\subsection{User-defined Extensions}

We proceed to briefly demonstrate how users can easily extend Lodestar by authoring custom blocks. All blocks in Lodestar inherit their core functionality from a \texttt{Block<TInputs..., TOutputs..., TParameters...>} class. There are two steps a user needs to take if they wish to add a new block: (1) inherit from the \texttt{Block<>} class with the desired paramters and define the new block's behavior, and (2) declare a specialization of the \texttt{BlockTraits} class. \texttt{BlockTraits} objects are used to determine whether a block has direct feedthrough, as well as the kind of block it is, and information about the inputs, outputs, and parameters of the block. If these two conditions are met, a user-defined block is treated as a first-class object in the Lodestar engine, allowing users to easily interface with custom software and hardware services.

We now turn to two practical applications involving the Lodestar engine.

\section{Applications}\label{sec:applications}

In this section, we discuss two applications of Lodestar: torque-control of robotic arm joints and an autofocus algorithm for single-view thermography.

\begin{figure}[t]
\centering
\includegraphics[width=0.8\linewidth]{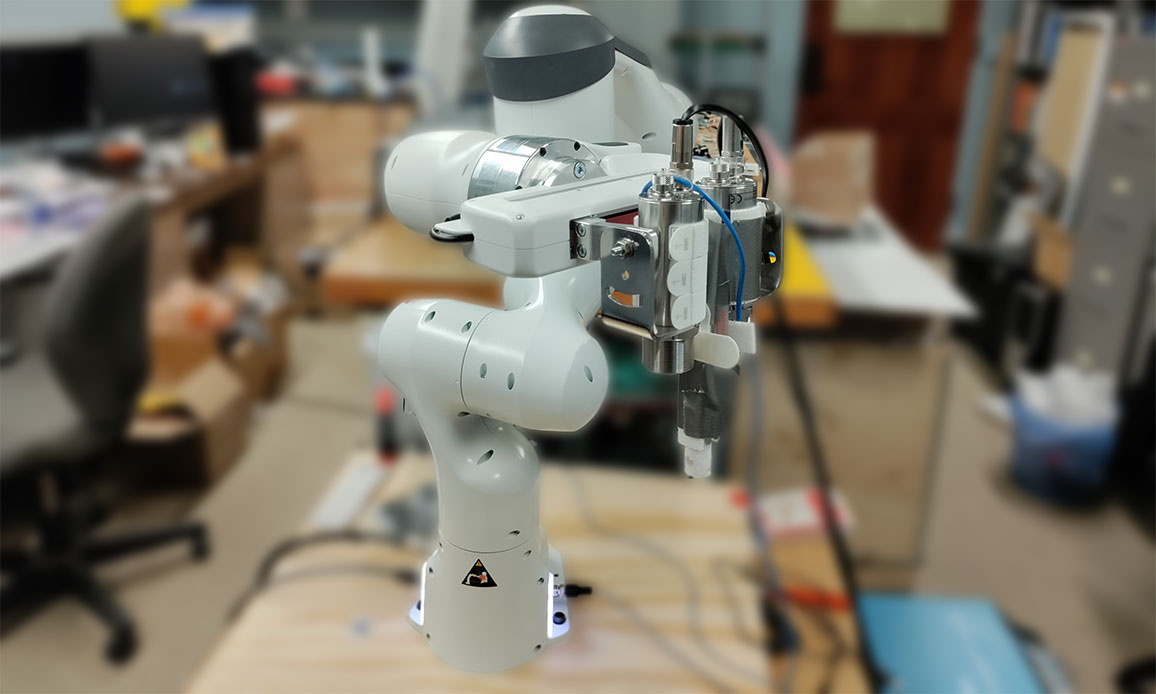}
\caption{Franka Emika Panda robotic arm and dual Optris Xi 400 microscopic thermographer setup attached to the gripper.}
\label{fig:hardware}
\end{figure}

\subsection{Robotic Arm Joint Torque Control}

We first consider the standard problem of joint configuration control using joint torques \cite{Cervantes2001}. Our goal is to show that one can easily construct a proportional--integral--derivative (PID) controller framework with anti-windup and saturation constraints in Lodestar. The classical and projection-based PID controllers are both applied in practice on a Franka Emika Panda 7-joint robotic arm (see Fig.~\ref{fig:hardware}), with real-time performance being compared. We also demonstrated that more advanced controllers can easily be prototyped using Lodestar, e.g., the fuzzy PID controller architecture of \cite{Malki1997}. We compare the performance of this latter controller in Lodestar to that of C++ code generated by Simulink Coder, with common use cases for either approach being discussed.

\begin{figure}[t]
\centering
\includegraphics[width=\linewidth]{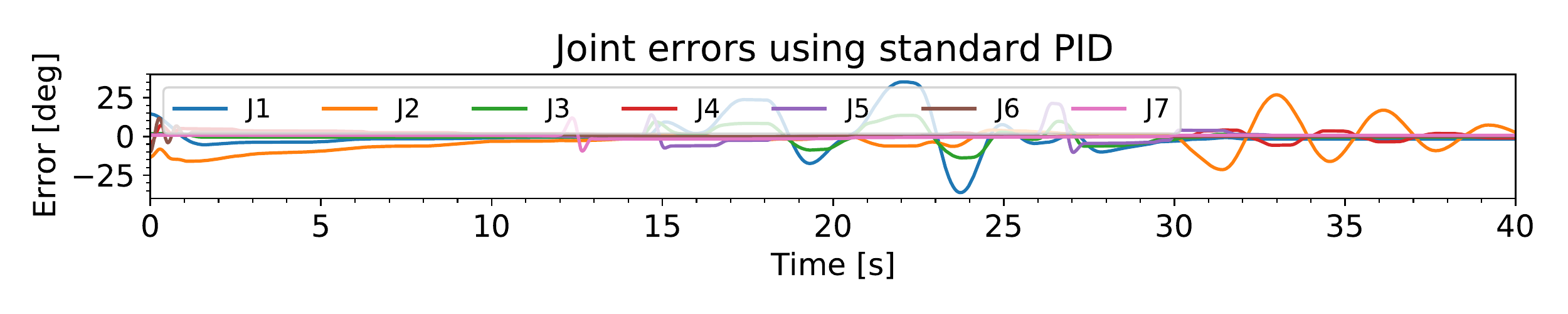}
\includegraphics[width=\linewidth]{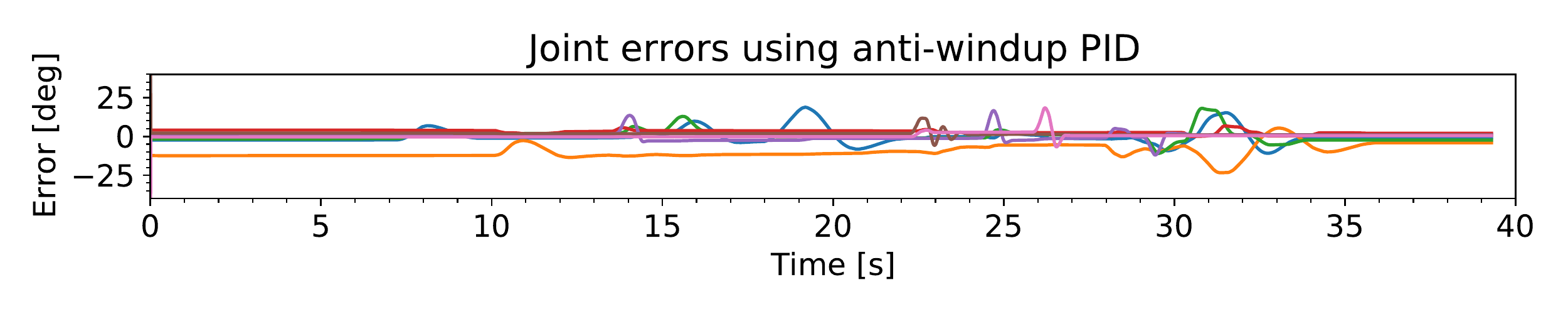}
\caption{Joint responses of classical and projection-based anti-windup PID, when subjected to external disturbances.}
\label{fig:pid-arm}
\end{figure}

We have implemented all algorithms on a workstation running Ubuntu 20.04 with a fully preemptible real-time kernel, equipped with 128 GB of RAM and a 24-core AMD Ryzen Threadripper 3960X. All code was compiled using GCC 11.2 using C++14 with \texttt{O3} optimization; the code is available on GitHub\footnote{\href{https://github.com/helkebir/Lodestar-Examples}{https://github.com/helkebir/Lodestar-Examples}}. A custom \texttt{PandaArmBlock} class was written to interface with the \texttt{libfranka} C++ drivers.
The results for the classical and projection-based anti-windup PID controllers are shown in Fig.~\ref{fig:pid-arm}.

Using the Franka Control Interface (FCI), a callback function may not take more than 1 millisecond for more than 5\% of all controller calls, imposing strict real-time constraints on the controller implementation. Using Lodestar, a single evaluation cycle for the classical PID controller takes an average of 12 microseconds with a standard deviation of 1 microsecond. This particular implementation uses the \texttt{SimplePIDBlock} class provided as part of Lodestar. To demonstrate the performance of custom controllers composed of multiple blocks, we have implemented the anti-windup PID controller using time delays, sums, gains, and saturation blocks. Despite the addition of separate blocks, we obtain an average cycle time of 32 microseconds with a standard deviation of 11 microseconds. This implies that there is over 930 microseconds of additional free processing time per torque callback in 99.7\% of the cases. Additionally, changes to the controller structure and gains can easily be made, allowing for rapid hardware-in-the-loop prototyping without relying on additional hardware or software, other than the compiler toolchain that would have been used to develop software for the target hardware.

To compare performance with generated code, such as provided by the Simulink Coder software, we have considered a single-input-single-output version of the fuzzy PID controller from \cite{Malki1997}. The generated code runs at 50 nanoseconds per cycle, while Lodestar code takes 390 nanoseconds on average. One should note, however, that any code generation solution needs to be accompanied with additional code for hardware interfacing, as well as custom hooks for interfacing with other software components; neither of these actions are needed when Lodestar is used. This makes Lodestar much more versatile than existing solutions, with rapid prototyping of different controller architectures becoming a possibility even on modest edge hardware. In addition, code generation takes at least two separate steps: (i) the model has to be \emph{transpiled} to source code, (ii) the generated source code has to be compiled and linked to auxiliary libraries. The former step already requires custom software that can often not run on modest hardware. Additionally, the latter step introduces additional vendor-specific code that is often undocumented. In Lodestar, all of the standard blocks are fully transparent and clearly defined in their respective header files, unlike packages that only provide binaries with no access to the original source files to identify potential compatibility issues.

We now turn to an application in real-time computer vision.

%

\subsection{Autofocus for Single-view Infrared Thermography}

\begin{figure}[t]
\centering
\includegraphics[width=\linewidth]{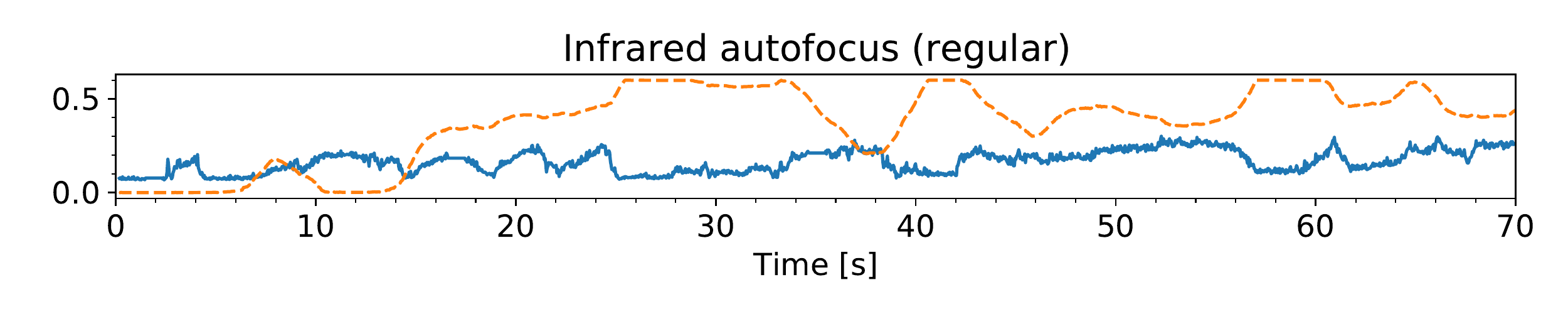}
\includegraphics[width=\linewidth]{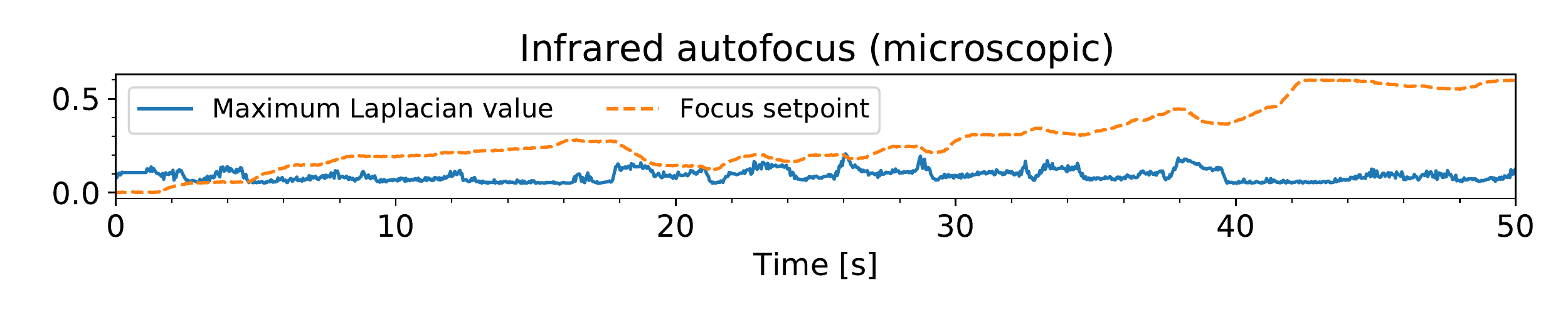}
\caption{Thermographic autofocus time series for two different sets of optics: mid-range optics (top), and microscopic optics (bottom). The solid blue line shows the maximum absolute Laplacian value, which is indicative of image sharpness. The orange dashed line is the focus motor set point, which saturates at 0.6. In response to changes in sharpness, our algorithm quickly refocuses in the correct direction.}
\label{fig:thermographer-timeseries}
\end{figure}

In energy-based surgery methods, thermal damage can often go unnoticed even by skilled physicians, often resulting in lasting damage that can only be detected days after a surgical procedure \cite{Palanker2008}. For this reason, our previous work proposes the use of infrared thermography to incorporate temperature information into sensing and control systems for robot-assisted surgery \cite{El-Kebir2021c, El-Kebir2021d}. For the purposes of system identification and control, careful calibration of the focus of a thermographer is essential to ensure that no artificial diffusion is introduced. Unlike in visible light imagery, image sharpness is hard to quantify in thermography \cite{Dziarski2021}. In this work, we present a novel algorithm based on noise-robust filtering to allow for focusing even in the presence of small temperature differences, as seen in electrosurgery \cite{El-Kebir2021d}. We briefly describe the algorithm below, showing its real-time performance in Lodestar when using two separate thermographers with different sets of optics (microscopic and mid-range).

\begin{figure}[t]
\centering
\includegraphics[width=\linewidth]{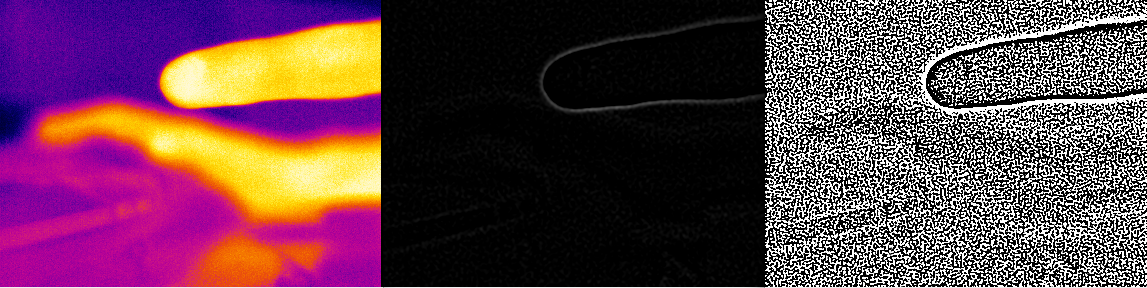}
\caption{Comparison of Laplacian filters when applied to thermographer imagery. On the left, the original thermal image is shown. The middle image shows the Laplacian obtained using a noise-robust filter; only the sharpest edges are clearly visible. A Sobel filter was used in the right image; high-frequency noise is strongly amplified.}
\label{fig:filter-comparison}
\end{figure}

Our algorithm relies on a noise-robust computation of the Laplacian of the image. The Laplacian has long been used as the backbone for edge detection in classical and modern computer vision alike, but a non-noise-robust version fails to perform satisfactorily in situations with small temperature gradients \cite{El-Kebir2022}. This effect can be seen in Fig.~\ref{fig:filter-comparison}, where the middle image was obtained using a noise-robust Laplacian filter, whereas a classical Sobel filter was used to create the right image; this latter image cannot be used for real-time autofocus, without application-specific thresholding. In \cite{El-Kebir2022}, the authors present a novel real-time parameter estimation method, known as attention-based noise-robust averaging (ANRA), applied to thermodynamic processes. In ANRA, the use of noise-robust convolution filters is key, since these filters accurately compute derivatives while suppressing high-frequency noise \cite{Holoborodko2009}.

After obtaining the Laplacian using this noise-robust filtering approach, we consider the maximum absolute value of the Laplacian field, which we call \emph{sharpness}. In particular, we take the error between the current sharpness value, and the one obtained 100 cycles ago (about 4 seconds when capturing images at 27 Hz) to render the algorithm aware of long-term changes. This long-term memory allows for direct determination of the focus direction without external aids. The buffered change in sharpness is then passed to a regular PID controller which determines the focus set point of the built-in motor. 

The thermographers used in our experiments are both part of the Optris Xi 400-series, one equipped with microscopic optics (focal length of 20 mm, field of view of $18^\circ \times 14^\circ$), and the other with medium-range optics (focal length of 12.7 mm, field of view of $29^\circ \times 22^\circ$). The algorithm was tested by tracking a finger in a noisy background containing reflecting metal surfaces; the results of these trials are shown in Fig.~\ref{fig:thermographer-timeseries}. Even in the case of microscopic thermography, in which edges are often harder to detect, our approach manages to rapidly adjust to changes in the overall image sharpness (see bottom plot). The same behavior is more pronounced in the case of mid-range thermography, as can be seen in the top plot in Fig.~\ref{fig:thermographer-timeseries}.

The algorithm described above was implemented in Lodestar and deployed on the same workstation. In terms of performance, this particular Lodestar application runs at around 26.92 Hz (limited by the thermographer refresh rate of 27 Hz). If we assume that the thermographer reports its results instantaneously at 27 Hz, we find an average processing time of 100 microseconds, with all processing times lying under 6 milliseconds in 99.7\% of the cases, leaving at least 31 milliseconds for subsequent processing tasks.

We now briefly consider how this algorithm could have been implemented in other frameworks. When working with ROS, while ROS bindings with the thermographer's C library do exist, they still require a separate ROS service to be written. Moreover, using Lodestar the entire internal state can be accessed by any other part of the program, whereas in ROS, each service would have to deliberately communicate its intermediate steps, which would require prior consideration during design.
 The introduction of new capabilities and communication channels can easily be achieved in Lodestar without changing the entire program structure.


\section{Conclusion}\label{sec:conclusion}

In this work, we have presented a novel control engine for real-time applications in applied robotics and control theory, named \emph{Lodestar}. We have identified a gap in currently available software packages for rapid controller implementation, with Lodestar providing a framework in which data-driven controller architectures can be modeled and directly executed on hardware, with little additional overhead. We have discussed several key features of Lodestar, and demonstrated its capabilities in two applications: one related to joint-level torque control of a robotic arm, and the other to real-time autofocus of single-view thermography. These applications serve to show that Lodestar provides a straightforward means to prototyping complex control systems, while still providing performant controller implementations.

In future work, we aim to focus on expanding Lodestar's capabilities to include formal verification methods, as well as code generation capabilities. In addition, we plan to develop a graphical user interface for interactive control system design. While the multi-disciplinary nature of Lodestar was hinted at in this work, we aim to fully leverage Lodestar's integrative capabilities to implement a self-contained monitoring and identification system for robot-assisted electrosurgery in the near future.

%

\bibliographystyle{IEEEtran}
\bibliography{LodestarPaper.bib}

\end{document}